\documentclass[final]{cvpr}
\usepackage{times}
\usepackage{epsfig}
\usepackage{graphicx}
\usepackage{amsmath}
\usepackage{amssymb}
\usepackage{subfigure}
\usepackage{float}
\usepackage{bm}
\usepackage[lined,boxed,commentsnumbered, ruled]{algorithm2e}
\usepackage[pagebackref=true,breaklinks=true,colorlinks,bookmarks=false]{hyperref}

\def\cvprPaperID{3300} % *** Enter the CVPR Paper ID here
\def\confYear{CVPR 2021}

\begin{document}

%%%%%%%%% TITLE
\title{PML: Progressive Margin Loss for Long-tailed Age Classification}

\author{\textit{Zongyong Deng}$^{1}$, \textit{Hao Liu}$^{1,2,}$\thanks{Corresponding author.}, \textit{Yaoxing Wang}$^{1}$, \textit{Chenyang Wang}$^{1}$, \textit{Zekuan Yu}$^{3}$, \textit{Xuehong Sun}$^{1,2}$\\
$^{1}$School of Information Engineering, Ningxia University, Yinchuan, China\\
$^{2}$Collaborative Innovation Center for Ningxia Big Data and Artificial Intelligence \\Co-founded by Ningxia Municipality and Ministry of Education, Yinchuan, China\\
$^{3}$Academy for Engineering and Technology, Fudan University, ShangHai, China\\
{\tt\small zongyongdeng\_nxu@outlook.com;
	liuhao@nxu.edu.cn; yaoxing.wang\_nxu@outlook.com}\\ {\tt\small chenyang.wang\_nxu@outlook.com;
	yzk@fudan.edu.cn;
	sunxh@nxu.edu.cn}	
}

\maketitle

%%%%%%%%% ABSTRACT
\begin{abstract}
In this paper, we propose a progressive margin loss~(PML)~approach for unconstrained facial age classification. Conventional methods make strong assumption on that each class owns adequate instances to outline its data distribution, likely leading to bias prediction where the training samples are sparse across age classes. Instead, our PML aims to adaptively refine the age label pattern by enforcing a couple of margins, which fully takes in the in-between discrepancy of the intra-class variance, inter-class variance and class center. Our PML typically incorporates with the ordinal margin and the variational margin, simultaneously plugging in the globally-tuned deep neural network paradigm. More specifically, the ordinal margin learns to exploit the correlated relationship of the real-world age labels. Accordingly, the variational margin is leveraged to minimize the influence of head classes that misleads the prediction of tailed samples. Moreover, our optimization carefully seeks a series of indicator curricula to achieve robust and efficient model training. Extensive experimental results on three face aging datasets demonstrate that our PML achieves compelling performance compared to state of the arts. Code will be made publicly.                 
\end{abstract}

%%%%%%%%% BODY TEXT
\section{Introduction}
\begin{figure*}[tb] 
	\begin{center}
		%\fbox{\rule{0pt}{2in} \rule{.9\linewidth}{0pt}}
		\includegraphics[width=1\linewidth]{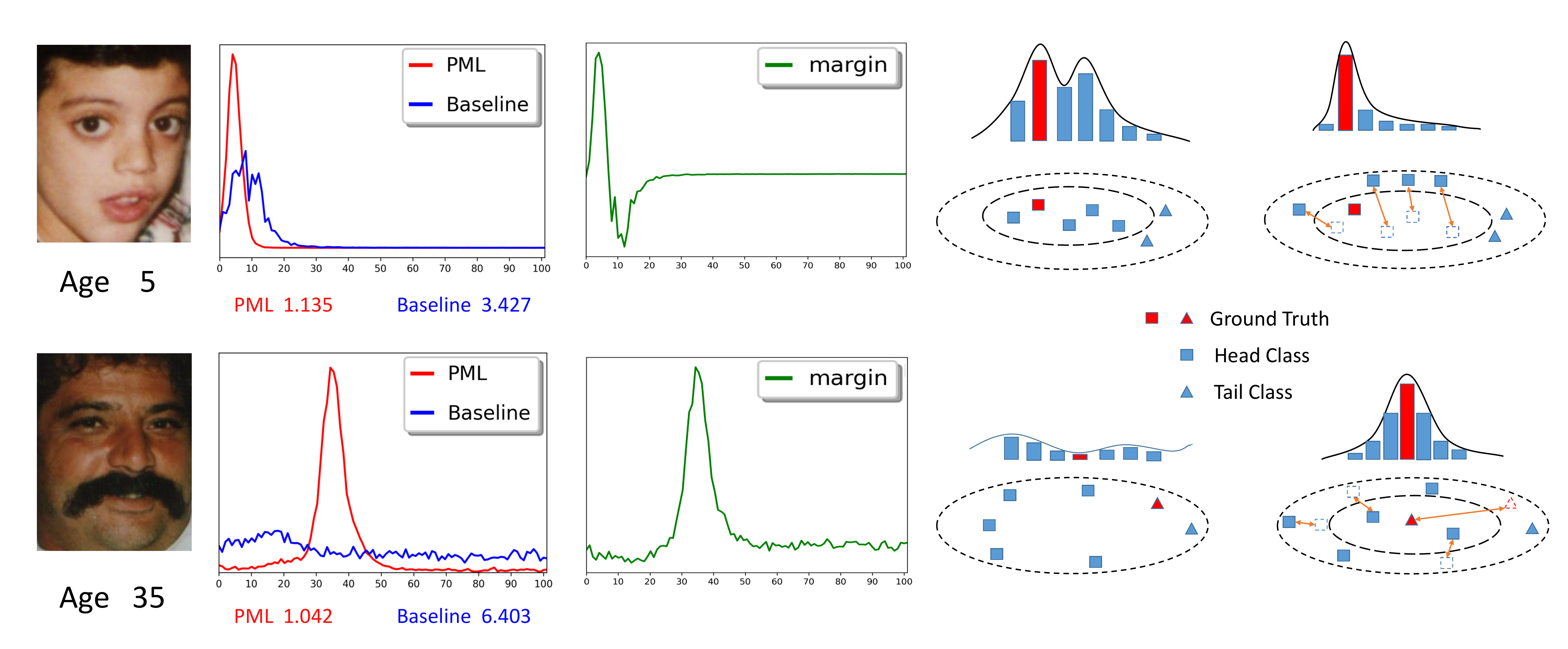}
	\end{center}
	\caption{Our approach versus existing label distribution learning approaches. We expect the ground truth~(red square or red triangle)~ to be at the center of the label prediction, where one sample in the head category~(blue square)~and another sample in the tail category~(blue triangle)~should be enforced by a margin from the real age. It is valuable to be notified that the dotted frame represents the position before adjustment. Top: The baseline method reasons the multi-modal distribution, because the head classes dominate the tail classes. Our proposed PML addresses this error by preventing the tail class from disturbance of the head. Bottom: The baseline method could hardly find effective features in tail categories and limited to output the uniform distribution. Our PML achieves robust feature representation by integrating the relation of adjacent age classes.~(Best viewed in color PDF file.)}
	\label{error}
\end{figure*}
%	\begin{figure}[tb] 
%	\begin{center}
%		%\fbox{\rule{0pt}{2in} \rule{.9\linewidth}{0pt}}
%		\subfigure[ChaLearn LAP 2015]{
%		\label{Fig1.1}
%		\includegraphics[width=0.45\linewidth, height=0.45\linewidth]{ChaLearn_distribution}}
%		\subfigure[FG-NET]{
%		\label{Fig1.2}
%		\includegraphics[width=0.45\linewidth,height=0.45\linewidth]{fg_distribution}}
%		\subfigure[One-hot bias]{
%		\label{Fig1.3}
%		\includegraphics[width=0.45\linewidth,height=0.45\linewidth]{one-hot}}
%		\subfigure[Multi-peak bias]{
%		\label{Fig1.4}
%		\includegraphics[width=0.45\linewidth,height=0.45\linewidth]{double_peak}}
%
%	\end{center}
%\caption{ Demonstration of problem when learning on long-tailed age data.~(a)~and~(b) represent the data distribution of ChaLearn LAP 2015 and FG-NET, respectively.~(c)~and~(d) take the class of age 14 in ChaLearn2015 as an example and indicate two common errors when training model in such minority class}\label{fig:insight}.
%\end{figure}
Facial age classification~(\textit{a.k.a.}, facial age estimation)~aims to predict the exact biological ages from given facial images, which has a lot of potential computer vision applications such as human-computer interaction~\cite{DBLP:journals/pami/ShuTLLZY18,DBLP:journals/nn/FragopanagosT05}~and facial attribute analysis~\cite{DBLP:journals/cg/MerillouG08,DBLP:journals/ejivp/AnguluTA18}. While numerous works have been devoted recently~\cite{DBLP:journals/tip/GaoXXWG17,li2019bridgenet,DBLP:conf/ijcai/GaoZWG18,rothe2015dex,pan2018mean}, the performance still remains limited in wild conditions, which is mainly due to that the datasets often undergo long-tailed distribution with many minority classes~(tail)~and a few common classes~(head). When learning with the long-tailed age data, a common problem is that the head classes usually dominate the training convergence. Therefore, the learned age classification model tends to perform better on head classes, whereas the performance degrades in tail classes. This quite motivates us to develop a robust facial age classification approach versus imbalanced age data. In the left of Fig.~\ref{error}, we visualize some failure cases caused by existing age classification methods. 

Facial age classification approaches could be roughly divided into the single label learning~(SLL)-based~\cite{DBLP:journals/tip/GaoXXWG17,DBLP:conf/ijcai/GaoZWG18,pan2018mean,deng2020learning} and  the label distribution learning~(LDL)-based~\cite{rothe2015dex,liu2017ordinal,2016Deep,DBLP:conf/cvpr/0002GWZWY18,DBLP:conf/iccvw/LiuLKZWLHSC15}. SLL-based methods typically classify one single age for a given facial image, which treats each age independently. However, they ignore human face changes gradually with progressive ages, thus the facial appearance is usually indiscriminative at adjacent age classes. To further model the age correlation,  Geng \textit{et al}.~\cite{geng2016label} proposed an LDL method to map the real-valued ground-truth to a Gaussian label distribution. However, the performance degrades in such long-tailed case where the feature representation of minority is suppressed by the majority classes.

To address the long-tailed data issue, we propose a progressive margin loss~(PML)~approach for age classification, which aims to leverage semantic margins to reduce intra-class variance and enlarge inter-class variance simultaneously. As shown in Fig.~\ref{fig:flowchart}, we carefully develop a progressive margin loss at the top of deep neural networks with preserving the age-difference cost information. Technically, our proposed PML is composed of two crucial branches including an ordinal margin learning and a variational margin learning. The ordinal margin attempts to extract discriminative features while maintaining the relation of the age order. In practice, the variational margin shifts the classifier decision boundaries of tail classes by transferring knowledge from the head classes. For efficient optimization, we develop a series of indicators by following the curriculum-learning method. To validate the effectiveness of our proposed method, we perform extensive experiments on three widely-used face aging datasets, where each dataset undergoes varying degrees of the imbalance. From the results, we achieve superior performance compared with the state-of-the-art methods especially only with fewer samples. For example, without using any external datasets, we decrease the MAE by 1.56 compared with the recently reported benchmark with only sparse and limited samples.
\section{Related Work}
In this section, we briefly review the  related works on facial age estimation and imbalanced classification, respectively.

\textbf{Facial Age Classification. }
Conventional age classification methods could be roughly divided into two types: feature representation~\cite{DBLP:journals/pami/AhonenHP06, DBLP:journals/pami/CootesET01, DBLP:conf/icip/EldibE10} and age prediction~\cite{DBLP:journals/tmm/FuH08, DBLP:conf/cvpr/ChangCH11}. Feature representation-based methods aim to exploit discriminative feature patterns from the facial images. Respectively, age prediction-based methods learn to classify the age labels with the extracted features. However, both types are optimized in a two-stage manner, which likely leads to local solution. To circumvent this limitation, deep learning has been applied to jointly optimize both procedures of feature representation and age prediction. For example, Rothe \textit{et al.}~\cite{rothe2015dex} formulated the age estimation as the expectation-based classification problem, where the prediction is accomplished by maximizing the expectation of outputting logits. Nevertheless, these methods hardly exploit the full chronological relationship of practical ages. To introduce the age label correlation to the model, Liu \textit{et al.}~\cite{liu2017ordinal} proposed an ordinal deep feature learning~(ODFL)~method, which enforces both the topology ordinal relation and the age-difference information in the learned feature space. Furthermore, Li \textit{et al.}~\cite{li2019bridgenet} developed the BridgeNet to model the ordinal relation of age labels via gated local regressors. To further alleviate the problem of label ambiguity, Geng \textit{et al.}~\cite{geng2016label} designed a distribution learning approach to transform the single scalar label to a vector. Nevertheless, the LDL schema gives rise to bias in predicting minority classes, where the samples within each age class are variant in appearance. We cope with this issue by a progressive margin loss framework, which elaborately adjusts the learned age patterns by fully considering the distributed property of neighboring age classes.   

\textbf{Imbalanced Classification. }
With the remarkable success achieved by data-driven CNNs~\cite{DBLP:conf/cvpr/LiLSBH15,DBLP:journals/pami/LiuLGWZ20,murphy2008head,geng2016label}, deep models have witnessed that the generation capacity is limited especially for imbalanced and distributed data~\cite{buda2018systematic,johnson2019survey}. Existing imbalanced classification methods are coarsely divided into re-sampling~\cite{ando2017deep,buda2018systematic,ren2017ensemble} and cost-sensitive loss function~\cite{lin2017focal,liu2019fair,DBLP:conf/cvpr/0002MZWGY19}. Accordingly, re-sampling-based methods aim to balance the scalability of the head classes and tail classes, but such schema easily prones to over-fitting in the tail classes. The major reason is that the training model memorizes irrelevant noise when utilizing the tailed data repeatedly~\cite{DBLP:conf/eccv/LiuCLL0LH20}. Cost-sensitive methods are developed to improve the influence of minority classes by assigning higher misclassification costs to the minority class than to the majority ones. In addition, both works~\cite{lin2017focal,liu2019adaptiveface} were proposed by the focal loss to mine hard-negative instances online and adaptive margin softmax to adjust the margins for different classes adaptively. However, these methods ignore the original relationship within samples \textit{w.r.t.} neighboring age classes. As far as one can tell from the literature, few works of imbalanced classification have been visited yet in facial age classification.

\section{Approach}
\begin{figure*}[tb] 
	\begin{center}
		%\fbox{\rule{0pt}{2in} \rule{.9\linewidth}{0pt}}
		\includegraphics[width=1\linewidth]{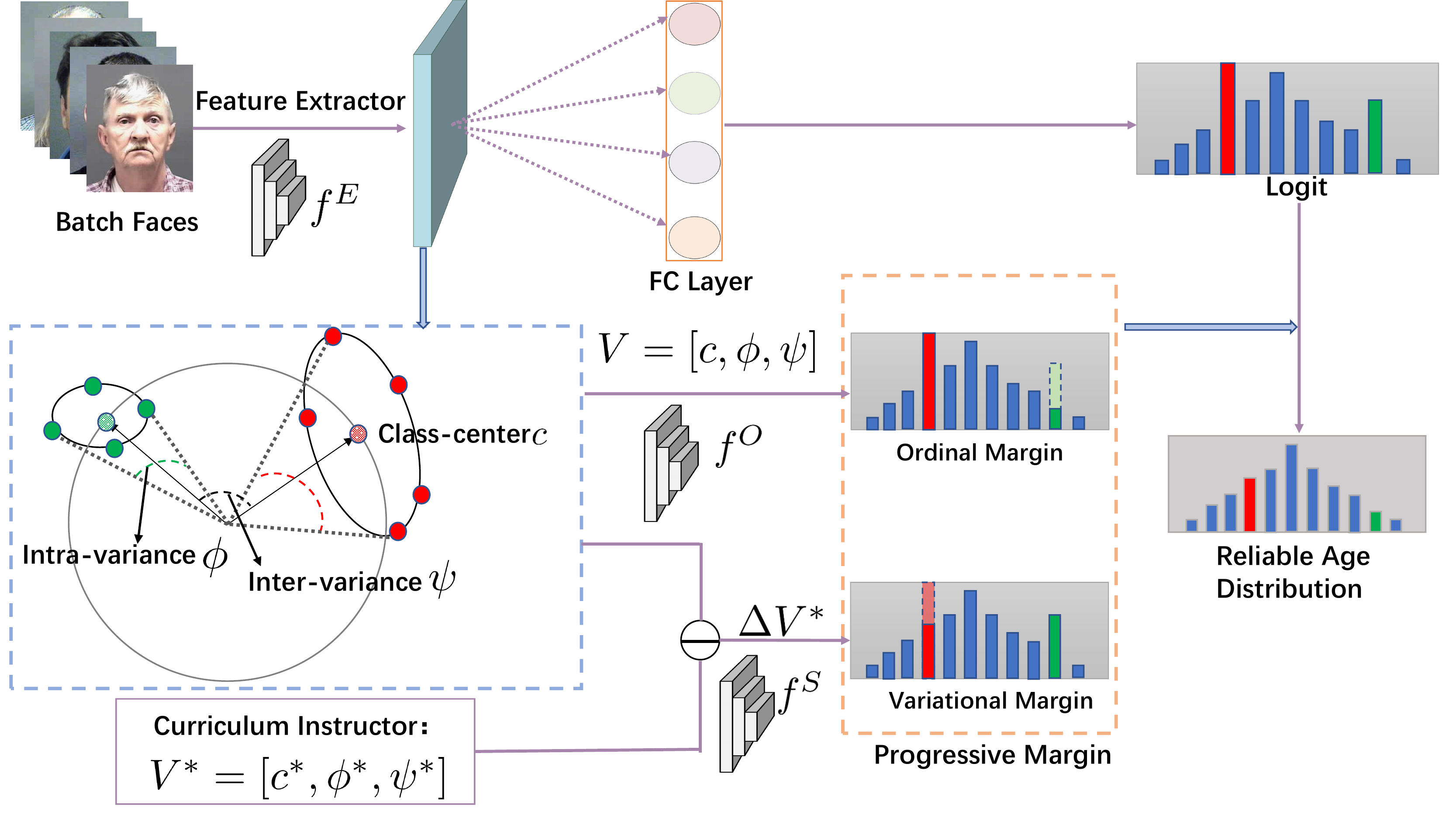}
	\end{center}
	\caption{ Flowchart of Our PML. Our architecture starts with the input faces, feeding to the feature extractor network $f^E(\cdot)$. Having obtained these deep features, we first compute class center $\bm{c}$, intra-class variance $\phi$ and inter-class variance $\bm{\psi}$ with each class as is shown in blue dotted rectangle. Then we reason out ordinal margin based on the concatenation of all variables~(\textit{i.e. $V$})~mentioned above. Meanwhile, we perform the residual $\Delta V^*$ by subtracting the preserved prior curriculum instructor variable $V^*$ from the concatenated variable $V$. Based on the residual, the variational margin is deduced. Finally, we fuse with the both types of progressive margins versus imbalanced age classes.}\label{fig:flowchart}
	
\end{figure*}
In this work, we claim that \textit{margin matters} for robust age label distribution learning. To achieve the proposed progressive margin loss, we enforce our model to integrate with the practical age progression in the learned distribution, which is semantic and interpretable. Fig.~\ref{fig:flowchart}  demonstrates the overall architecture of our proposed method. In detail, the PML contains three components: a backbone feature extractor $f^E(\cdot)$, an ordinal margin learning branch $f^O(\cdot)$ and a variational margin learning branch $f^V(\cdot)$. For an input image $I$ , the feature denoted by $\bm{x}$ is extracted by the layer-4 of the backbone ResNet-34 network~\cite{DBLP:conf/cvpr/HeZRS16}. Then we define the class center $\bm{c}$, the inter-variance $\phi$ and the intra-variance $\bm{\psi}$, which will be updated according to the recursive formula for calculating the mean and the variance. Moreover, our approach learns both the ordinal and variational margins by taking $\bm{c}$, $\phi$ and $\bm{\psi}$ as the inputs to the $f^O(\cdot)$ and $f^V(\cdot)$.  Finally, we introduce a curriculum learning protocol~\cite{DBLP:conf/aaai/JiangMZSH15,DBLP:conf/icml/GravesBMMK17} to smoothly simulate data distribution from being balanced to imbalanced. To clarify the notations, Table~\ref{tab:symbol} tabulates the detailed descriptions of all employed variables and functions in this work.

\subsection{Problem Formulation}
Let $y \in \{0, ..., 100\}$ denote the ground-truth age for each input $I$. Based on the property of label ambiguity, a facial image feature responds to different similarities across ages and the similarity roughly obeys the Gaussian distribution~\cite{DBLP:journals/tip/GaoXXWG17}. Our goal is to transform the scalar age value $y$ to an adaptive label distribution $\bm{y} \in \mathbb{R}^{101}$ as follows.
\begin{equation}
{y}_k=\frac{1}{\sigma\sqrt{2\pi}}exp(-\frac{(k-y)^2}{2\sigma^2}),
\end{equation}
where $k\in [0, 100]$, $\sigma$ is the variance of label distribution. ${y}_k$ is the $k$-th element of $\bm{y}$ which represents the probability that the true age is $k$ years old, respectively.

To classify the progressive ages with the long-tailed data, we propose a progressive margin loss to reason out robust label distributions. To this end, our approach maintains the ordinal age correlation and suppresses the noise of majority classes on the minority ones in the learned feature space. Moreover, our approach leverages Kullback-Leibler(KL) divergence to measure the distance between the ground-truth distribution and the predicted one. 

\begin{table}[tb] \small
	\caption{Detailed description of the variables.
	}
	\centering
	\label{tab:symbol}
	\vspace{6pt}
	\renewcommand\arraystretch{1.2} 
	\begin{tabular}{|c|l|}
		
		\hline \textbf{Symbol} & \textbf{Definition} \\
		\hline
		
		$I \in \mathbb{R}^{W \times H}$ & Raw face image with  ${W \times H}$ pixels\\
		$\bm{x} \in \mathbb{R}^{D}$ & Extracted feature with $D$-dimension \\
		$\bm{y} \in \mathbb{R}^{c}$ & Age label distribution consisting \\ 
		& of $c$ Age scalar values $y$\\
		$\bm{c} \in \mathbb{R}^{c \times D}$ &  Class centers with $D$-dimension \\
		$\phi \in \mathbb{R}^{c \times 1}$ &   Intra-class variances \\
		$\bm{\psi}  \in \mathbb{R}^{c \times c}$&  Inter-class variances  \\
		$V \in \mathbb{R}^{c\times (D+1+c)} $&  Concatenation of the $\bm{c}$, $\phi$ and $\bm{\psi}$ \\	
		$s(\cdot)$ &  Dot similarity measure  function \\
		$d(\cdot)$ &  Cosine distance measure function \\ 
		$f^O(\cdot)$ &  Function of ordinal margin learning  \\
		$f^V(\cdot)$ &  Function of variational margin learning  \\
		$M_o \in \mathbb{R}^{c \times 2}$ &  Ordinal margins including a tuple of \\ 
		& mean and variance \\ 
		$M_v \in \mathbb{R}^{c \times c}$ &  Variational margins computing by\\
		&  one-vs.-all~(OvA) schema\\ 
		
		\hline
	\end{tabular}%
\end{table}
In this way, the optimal parameter $\theta^*$ is determined by
\begin{equation}
\begin{aligned}
\theta^*&=\mathop{\arg\min}\limits_{\theta}\frac{1}{n}\sum_{i=1}^{n}\bm{y}_{i}log\frac{\bm{y}_{i}}{\bm{\hat{y}}_i} \\&=\mathop{\arg\min}\limits_{\theta}-\frac{1}{n}\sum_{i=1}^{n}\bm{y}_{i}log\bm{\hat{y}}_i,\label{kl}
\end{aligned}
\end{equation}
	     
Actually, Equ.\ref{kl} is the softmax cross-entropy loss function, which was widely used in the margin-based metric learning~\cite{DBLP:conf/iccv/HayatKZS019,cao2019learning,liu2019fair}. The main insight of these methods is to enforce the intra-class concentrations and inter-class diversity by introducing margins to the softmax loss. However, these methods only consider single labels independently, thus ignoring correlated information of neighboring ages. Hence, the fixed positive margin is inflexible to exploit the real-world age distribution. To address the aforementioned problem, our PML suits the distribution learning framework by the newly-learned margins and moreover can be optimized by the standard back-propagation algorithm. The PML is formulated as follows.
\begin{equation}
\mathcal{L}_{m_p} =-\frac{1}{n}\sum_{i=1}^{n}\bm{y}_{i}log\bm{\hat{y}^*}_i, \label{loss mp}
\end{equation}    
\begin{equation}
\begin{aligned}
\bm{\hat{y}^*}_i =[\frac{exp(s(\bm{x}_i,W_1)-m_{p1})}{exp(s(\bm{x}_i,W_1)-m_{p1})+\sum_{t \neq 1}exp(s(\bm{x}_i,W_t))},
\\...,\frac{exp(s(\bm{x}_i,W_c)-m_{pc})}{exp(s(\bm{x}_i,W_c)-m_{pc})+\sum_{t \neq c}exp(s(\bm{x}_i,W_t))}]^\mathrm{T}\label{margin p},
\end{aligned}
\end{equation}         
where $s(\cdot)$ denotes the similarity function, \textit{e.g.} dot product similarity~\cite{DBLP:conf/icml/LiuWYY16}, and $m$ denotes the parameters for our learned margins of the $k$-th class, respectively. 

Obviously, how to learn the appropriate and interpretable margins is a crucial part in our PML. Since long-tailed age classification is determined by the chronological relation and imbalanced degree of data simultaneously, our proposed PML takes both factors into account in our learned margins, which could be optimized in a globally-tuned manner.        

\subsection{Progressive Margin Loss}
The proposed progressive margin loss framework mainly includes the ordinal margin learning module and the variational margin learning module. To be specific, the ordinal margin aims at making the deep feature more discriminative and simultaneously preserving the ordinal correlation. We assume that the class center is the mean of its samples, which responses the holistic property of one class in the embedding space. In other words, it not only indicates the feature discriminability, but also includes the discrepancy between the majority and minority class. Since a high-level feature $\bm{x}$ embeds abundant semantic information of the input sample, we take $\bm{x}$ to represent this sample. For each $\bm{x}$ \textit{w.r.t} one class, the class center is performed as follows.
\begin{equation}
\bm{c}_j = \frac{1}{N_j}\sum_{i=1}^{N_j}\bm{x}_i,\label{class center}
\end{equation}  
where $\bm{c_j}$ and $N_j$ denote the center of $j$-th class and the number of samples in the $j$-th class, and $\bm{x}_i$ is the feature which belongs to this class, respectively. However, Equ.\ref{class center} requires $N_j$ instances of $j$-th class~(\textit{i.e.} the whole instances belong to $j$-th class), which cannot be directly applied to mini-batch iterative training. Based on the recursive formula for calculating the mean, our PML computes the class center as follows.  
\begin{equation}
\bm{c}_j^{t} = \bm{c}_j^{t-1} + I(y_i=j)\frac{I(y_i=j)\bm{x}_i-\bm{c}_j^{t-1}}{N_j^{t-1}+1},\label{update C}
\end{equation}
where $t$ denotes the training iterations and $I(\cdot) \in \{0,1\}$ is the indicator function. $I(\cdot)$ outputs 1 only if the conditions in brackets are true, vice versa. According to the class center $\textbf{c}$, the inter-class variance is performed as follows.
\begin{equation}
\bm{\psi}_j^t = [d(\bm{c}_j^t,\bm{c}_0), ..., d(\bm{c}_j^t,\bm{c}_c)],\label{inter-class}
\end{equation}
where $d(\cdot)$ denotes the cosine distance measure function~\cite{DBLP:conf/accv/NguyenB10}. 

To compute the intra-class, we reformulate the recursive formula for calculating the variance as follows.
\begin{equation}
\phi_j^{t} = \phi_j^{t-1} + I(y_i=j)d(\bm{x}_i, \bm{c}_j^{t-1})d(\bm{x}_i - \bm{c}_j^{t}),\label{intra-class}
\end{equation}   
where $\bm{c}_j$, $\phi_j$ and $\bm{\psi}_j$ response holistic feature representation, intra-class variance and inter-class variance of $j$-th class, respectively. 

Our proposed PML concats all of them as inputs to $f^O(\cdot)$ to get ordinal margin. For simplicity, we assume that $M_o$ obeys the Gaussian distribution. Having enforced the constraint to our ordinal margin network, the margins are generated as below.
\begin{equation}
M_o = f^O([\bm{c},\phi,\bm{\psi}]), M_o \in \mathbb{R}^{c\times 2},\label{M_o}
\end{equation} 
where $[\bm{c},\phi,\bm{\psi}] \in \mathbb{R}^{c\times (D+1+c)}$ denotes the concatenation of these variables. Note that $M_o$ is composed of the computed mean and variance, which transforms to $M_o^* \in \mathbb{R}^{c \times c}$ by discretely sampling from the range of 0 to $c$. After combing Equ.\ref{M_o} with Equ.\ref{loss mp}, the ordinal margin can be optimized in a unified framework.     

$M_o$ enhances the feature discriminativeness by considering the age ordinal relation. However, it may fall into a sub-optimal solution for adjacent age classes with imbalanced training samples. In such case, the minority age samples are likely misclassified to the majority age labels. Instead, our proposed variational margin is progressively suppressed the majority classes with its own influence. We performed the residual of class center, inter-class variance and intra-class variance between to adjacent iterations as.
\[
\Delta V = [\bm{c}^t,\phi^t,\bm{\psi}^t]-[\bm{c}^{t-1},\phi^{t-1},\bm{\psi}^{t-1}],
\]
\begin{equation}
M_v = f^V(\Delta V), M_v \in \mathbb{R}^{c}. \label{M_v}
\end{equation}          

Noticing that $M_o^* \in \mathbb{R}^{c \times c}$ exploits the relation about the one-vs.-all~(OvA) mechanism~\cite{bishop2006pattern}, which pays attention to local examples. $M_v \in \mathbb{R}^c$ is complementary to $M_o^*$ by enhancing the learned feature of each class especially for the minority class, which efficiently prevents the disturbance of other classes. We formulate this sense as follows.
\begin{equation}
M_{pj} = \lambda M_{o}^* + \beta M_v,\label{M_p}
\end{equation}  
where $M_{pj}$ denotes progressive margin, $\lambda$ and $\beta$ is used to balance $M_o^*$ and $M_v$, respectively. 

\begin{algorithm}[tb] \label{alg:pml}
	\DontPrintSemicolon
	\LinesNumbered
	\KwIn{Training set: $\mathcal{D}=\{I_i\}^{i=1:n}$, maximal iteration $T$.}
	\KwOut{Parameters of $f^E(\cdot)$, $f^O(\cdot)$ and $f^V(\cdot)$  .}
	\For{$ t < T$}{
		/*Extracting the feature of $i$-th face image.*/ \\
		$\bm{x}_i = f^E(I_i)$; \\	
		/*Assuming $\bm{x}_i$ belongs to $j$-th class and updating the class of each class by Equ.\ref{class center}.*/ \\
		$\bm{c}_j^t = \bm{x}_i$, \quad /*For iteration-1*/; \\
		$\bm{c}_j^t = \bm{c}_j^{t-1}+\frac{\bm{x}_i-\bm{c}_j^{t-1}}{N_j^{t-1}}$; \\
		
		/*Updating the inter-class variance by Equ.\ref{inter-class}.*/\\
		$\bm{\psi}_j^t = [d(\bm{c}_j^t,\bm{c}_0), ..., d(\bm{c}_j^t,\bm{c}_c)]$; \\
		/*Updating the intra-class variance by Equ.\ref{intra-class}.*/\\
		$\phi_j^{t} = \phi_j^{t-1} + d(\bm{x}_i - \bm{c}_j^{t-1})d(\bm{x}_i - \bm{c}_j^{t})$;\\
		/*Learning the ordinal margin by Equ.\ref{M_o}.*/\\
		$M_o = f^O([\bm{c},\phi,\bm{\psi}])$;\\
		/*Optimization with Curricula.*/
		$\Delta V = [\bm{c}^t,\phi^t,\bm{\psi}^t]-[\bm{c}^{*},\phi^{*},\bm{\psi}^{*}]$;\\
		/*Learning the variational margin by Equ\ref{M_v}.*/\\
		$M_v = f^V(\Delta V), M_v \in \mathbb{R}^{c}$;\\
		/*Optimizing our PML by Equ.\ref{M_p} and Equ.\ref{loss mp}.*/\\
		$M_{pj} = \lambda M_{o}^* + \beta M_v$;\\
		$\mathcal{L}_{m_p} =-\bm{y}_{i}log\bm{\hat{y}^*}_i$;\\
		
	}
	\textbf{Return}: $f_\theta^E(\cdot)$, $f_\theta^O(\cdot)$ and $f_\theta^V(\cdot)$.
	\caption{Training Procedure of Our PML}
\end{algorithm}

\subsection{Optimization with Curricula}
To further make the margin learning process more stable and fast, our proposed PML follows the insight of curriculum learning~\cite{DBLP:conf/aaai/JiangMZSH15,DBLP:conf/icml/GravesBMMK17}. Specifically, we divide the training data into five curricula to optimize the network parameters, where each curriculum consists of varying degrees of data imbalance. In this way, the proposed PML is learned by gradually including samples distribution from being balanced to imbalanced. Unlike classic curriculum learning mechanism where each curriculum contains non-crossing label fields, we design a sampling method to model the consistency of label fields. Our protocol of curriculum learning is defined as
\[
\mathcal{D}_1 \subset \mathcal{D}_2 \subset \mathcal{D}_3 \subset \mathcal{D}_4 \subset \mathcal{D}_5, \quad \mathcal{D}_5=\mathcal{D}_{all},
\]
\begin{equation}
\begin{aligned}
\mathcal{D}_{i}&=\left\{X_{i}, Y_{i}\right\}, \\
{s.t.} \quad\quad X_{i}&=x^{\left(0, \delta_{i}\right)} \cup  \bm{\rho} \left(x^{\left(\delta_{i}+1, c\right)}\right).\label{eq.curriculum}
\end{aligned}
\end{equation} 
For splitting, we firstly sort each class by its owned instances in an ascending order, where $\delta_{i}$ denotes the dividing line of the $i$-th course. Then, the function of $\bm{\rho}(x^{(a,b)})$ represents a sampling operation, which draws the same number of instances as the $(a-1)$-th class from the range of $a$ to $b$.       

More specifically, our PML takes data from $D_1$ to $D_5$ as the inputs to train the deep convolution neural networks,  until it converges in each curriculum. Based on the class property of previous curriculum $V_{pre}=[\bm{c}^*,\phi^*,\bm{\psi}^*]$ , the variational margin between the adjacent curricula can be obtained as
\[
\Delta V^* = [\bm{c}^t,\phi^t,\bm{\psi}^t]-V_{pre},
\]
\begin{equation}
M_v^* = f^V(\Delta V^*), M_v \in \mathbb{R}^{c}. \label{M_v*}
\end{equation}   

Since the class property $V_{pre}$ is acquired in a further balanced course than current ones, the learned $V_{pre}$ is the unbiased representation towards each class. By referencing this unbiased instructor, the learning procedure of $M_p$ becomes stable. Through this curriculum learning fashion, the optimization process is slightly affected by data imbalance. Experimentally, we find out that this learning schema can achieve comparable performance with the state-of-the-art methods by using fewer training samples.  
Note that we only enforce theses margins in the training procedure for achieving discriminative feature representation. 

\textbf{Algorithm}~\ref{alg:pml} shows the optimization procedure of the proposed PML.

\begin{figure*}[tb] 
	\begin{center}
		%\fbox{\rule{0pt}{2in} \rule{.9\linewidth}{0pt}}
		
		\includegraphics[width=1\linewidth]{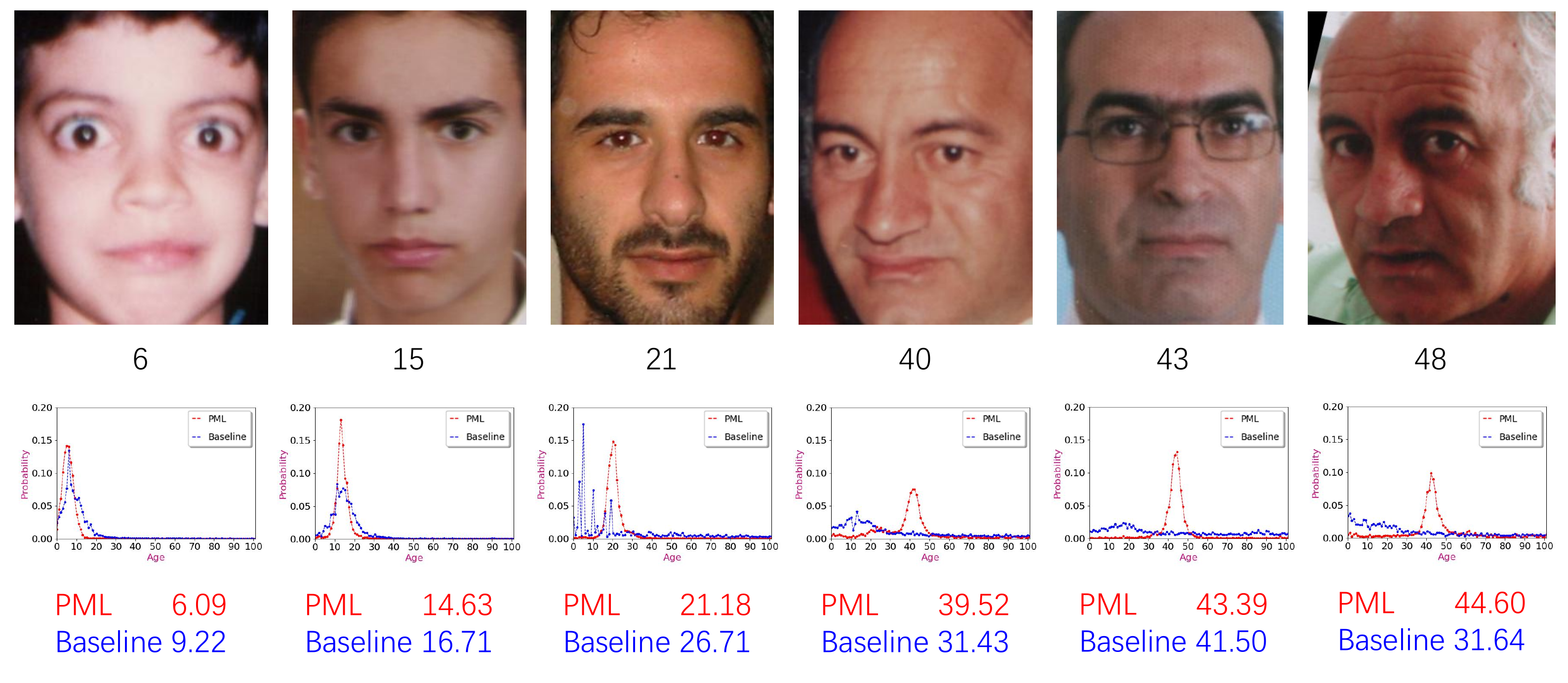}
	\end{center}
	\caption{Comparisons of predicted distributions on FG-NET. The first row shows six aligned faces and their corresponding ground-truths. The second row shows the predicted distributions of the baseline and our PML approach. Seeing from these distributions, in our method, the predictions is more accurate and reliable than baseline method. }\label{fig:predict}
	\label{distribution}
\end{figure*}

\section{Experiments}

To evaluate the effectiveness of the proposed method, we conducted experiments on three widely-used datasets for uncontrolled age classification. We conducted experiments our method on Morph II~\cite{2006MORPH}, FG-NET~\cite{2016Overview} and ChaLearn LAP 2015~\cite{DBLP:conf/iccvw/EscaleraFPBGEMS15}. For fair comparisons, we only used additional IMDB-WIKI dataset~\cite{2016Deep} for pre-training to evaluate ChaLearn LAP 2015.

\subsection{Evaluation Datasets and Metrics}

\textbf{Evaluation Datasets.} \textit{Morph II}. This database is the widely-used benchmark for age estimation, which consists of 55,134 face images of 13,617 subjects. The age range of this database covers from 16 to 77 years old. In our experiments, we used two types of testing protocols in our evaluations.~\textbf{Setting I.}~The dataset was randomly divided to training part~(80\%)~and testing part~(20\%).~\textbf{Setting II.}~A subset of 5,493 face images from Caucasian descent followed the work~\cite{DBLP:journals/pami/TanWLZGL18}.

\textit{FG-NET}. The FG-NET dataset contains 1,002 face images of 82 subjects and the age ranges from 0 to 69. We followed the previous methods~\cite{liu2017ordinal,pan2018mean} to use leave-one-out~(LOPO)~setting for evaluation.

\textit{ChaLearn LAP 2015}. This dataset was released in 2015 at the ChaLearn  LAP challenge, which collects 4,691 images. The ChaLearn LAP was labeled with the apparent age, and each label was set as an average of at least 10 people. This dataset contains training, validation and testing subsets with 2476, 1136 and 1079 images, respectively.

\textit{IMDB-WIKI}. The IMDB-WIKI consists of 523,051 images in total and the range is from 0 to 100. To follow the common setting, we selected about 300,000 images for training, where all non-face and severely occluded images were removed.

\textbf{Evaluation Metrics.} In the experiments we leveraged Mean Absolute Error~(MAE) to calculate the discrepancy between estimated age and the ground-truth. Obviously, the lower the MAE value, the better performance it achieves. According to previous work~\cite{DBLP:conf/iccvw/LiuLKZWLHSC15}, we also used the $\epsilon$-error to measure the performance on the ChaLearn dataset. In particular, this standard testing protocol is defined as follows. 
\[
\epsilon = 1 - \sum_{i=1}^{n}exp(-\frac{(y_i-y_i^*)^2}{2{\sigma_i^*}^2}),
\]
where $y_i^*$ is the ground-truth age value, $\sigma^*$ is the annotated standard deviation, respectively.    

\subsection{Implementation Details}

For each input image, we first detected the whole face with MTCNN~\cite{DBLP:journals/spl/ZhangZLQ16}. Then we aligned it based on the detected facial landmarks. For IMDB-WIKI, we straightly removed invalid images. In the training stage, we augmented all images randomly with horizontal flipping, scaling, rotation and translation. Moreover, we adopted ResNet-34~\cite{DBLP:conf/cvpr/HeZRS16} as our backbone network and this network was pretrained on ImageNet~\cite{DBLP:conf/cvpr/DengDSLL009}. For all experiments, we employed the Adam optimizer and SGD optimizer~\cite{DBLP:journals/nn/Qian99}. The weight decay and the momentum were set to 0.0005 and 0.9, respectively. The initial learning rate was set to 0.0001 and we leveraged two methods for learning rate adjustment. $\lambda$ and $\beta$ were tuned by cross validations. In the Adam optimization method, we used CosineAnnealingLR~\cite{DBLP:conf/iclr/LoshchilovH17} to adjust learning rates. Meanwhile, we used ExponentialLR for the SGD optimizer. For parallel acceleration, we trained our model with PyTorch~\cite{paszke2017automatic} on 4 Tesla V100 GPUS.

\subsection{Results and Analysis}

\textbf{Comparisons on Morph II.}~Table~\ref{tab:morph1} and Table~\ref{tab:morph2} show the MAEs of our approach on Morph II dataset with different settings. Noticing that we did not use IMDB-WIKI for pretraining in this dataset. According to the results, our model achieves 2.150 and 2.307 under the Setting I and Setting II, respectively. More specifically, in Setting I, our method achieves the best performance among all models except AVDL, but this model was pretrained on IMDB-WIKI. In Setting II, our model achieves the best performance among all state-of-the-art methods regardless of using the external datasets. From the results, we made two-fold conclusions:~(1)~Compared label distribution learning methods such as DLDL-V2~\cite{DBLP:conf/ijcai/GaoZWG18} and M-V Loss\cite{pan2018mean} leverages a fixed pattern to learned feature. Such schema ignores the issue of age imbalance, which likely hurts the discriminativeness of minority features.
~(2)~Particularly from the results on Setting II, we see that our PML outperforms most state of the arts with sparse training data. This achievement is due to that the learned margin enlarges the inter-class variance by preserving the age-related semantic information.    
\begin{table}[tb] \small
	\caption{Comparisons of MAEs of our approach compared with different state-of-the-art methods on Morph II under Setting I. Bold indicates the best~($^\ast$ indicates the model was pre-trained on the IMDB-WIKI dataset and $^\dagger$ indicates the model was pre-trained on the  MS-Celeb-1M, respectively. We annotated the 2nd performance as the italic type.~)}
	\vspace{5pt}
	\label{tab:morph1}
	\centering
	{
		\begin{tabular}{|c|c|c|}
			\hline
			\textbf{Method} & \textbf{Morph} &\textbf{Year}  \\
			\hline
			\hline
			OR-CNN~\cite{DBLP:conf/cvpr/NiuZWGH16} & 3.34 & 2016 \\		
			ODFL~\cite{liu2017ordinal} & 3.12 & 2017 \\
			ARN~\cite{DBLP:conf/iccv/AgustssonTG17} & 3.00 & 2017 \\	
			CasCNN~\cite{DBLP:journals/tcyb/WanTLGL18} & 3.30 & 2018 \\
			M-V Loss\cite{pan2018mean} & 2.41/2.16$^\ast$ & 2018 \\
			DRFs~\cite{DBLP:conf/cvpr/0002GWZWY18} & 2.17 & 2018 \\
			DLDL-V2~\cite{DBLP:conf/ijcai/GaoZWG18}  & 1.97$^\dagger$ & 2018 \\
			
			SADAL~\cite{DBLP:journals/tmm/LiuSZWYS20} & 2.75 & 2019 \\
			BridgeNet & 2.38$^\ast$ & 2019 \\
			AVDL~\cite{DBLP:conf/cvpr/ZhangLXZ19} & \textbf{1.94}$^\ast$ & 2020\\ 
			
			\hline
			\textbf{PML} & \textit{2.15}  &  \textbf{-} \\ 
			\hline
		\end{tabular}
	}
\end{table}
 
\begin{table}[tb] \small
	\caption{Comparisons of MAEs of our approach compared with different state-of-the-art methods on Morph II under Setting II.}
	\centering
	\label{tab:morph2}
	\vspace{5pt}
	\begin{tabular}{|c|c|c|}
		\hline
		\textbf{Method} & \textbf{Morph} &\textbf{Year}  \\
		\hline
		\hline
		DEX~\cite{rothe2015dex} & 3.25/2.68$^\ast$ & 2018 \\
		AgeED~\cite{DBLP:journals/pami/TanWLZGL18} & 2.93/2.52$^\ast$ & 2018 \\
		DRFss~\cite{DBLP:conf/cvpr/0002GWZWY18} & 2.91 & 2018 \\
		DHAA~\cite{DBLP:conf/ijcai/TanY0GL19} & 2.49$^\ast$ & 2019 \\
		AVDL~\cite{DBLP:conf/cvpr/ZhangLXZ19} & \textit{2.37}$^\ast$ & 2020 \\
		\hline
		\textbf{PML} & \textbf{2.31} & \textbf{-}	\\
		\hline
	\end{tabular}%
\end{table}

\textbf{Comparisons on FG-NET. }As shown in Table~\ref{tab:fg}, we compared our model with the state-of-the-art models on FG-NET. Our method PML achieves the lowest MAE of 2.17. Moreover, compared with AVDL that was pretrained by IMDB-WIKI, our PML decreases the MAE by 0.17 with our progressive margin loss. Compared with the state-of-the-art DHAA that was trained from scratch, our PML decreases the MAE by a large margin. Obviously, the results show that our method significantly works well on few-shot dataset.
\begin{table}[tb] \small
	\caption{Comparisons of MAEs of our approach compared with different state-of-the-art methods on the FG-NET dataset.}
	\centering
	\label{tab:fg}
	\vspace{5pt}
	\begin{tabular}{|c|c|c|}
		\hline
		\textbf{Method}  & \textbf{FG-NET} & \textbf{Year} \\ 
		\hline
		\hline
		DEX~\cite{rothe2015dex} & 4.63/3.09$^\ast$   & 2018 \\
		DRFs~\cite{DBLP:conf/cvpr/0002GWZWY18}       & 3.85   & 2018 \\
		M-V Loss~\cite{pan2018mean}   & 4.10/2.68$^\ast$   & 2018 \\
		AgeED~\cite{DBLP:journals/pami/TanWLZGL18}  & 4.34/2.96$^\ast$   & 2018 \\
		C3AE~\cite{DBLP:conf/cvpr/ZhangLXZ19}       & 2.95   & 2019 \\
		BridgeNet~\cite{li2019bridgenet}  & 2.56$^\ast$   & 2019 \\
		DHAA~\cite{DBLP:conf/ijcai/TanY0GL19}      & 3.72/2.59$^\ast$   & 2019 \\
		AVDL~\cite{DBLP:conf/cvpr/ZhangLXZ19}     & \textit{2.32}$^\ast$    & 2020 \\
		NRLD~\cite{deng2020learning}        & 2.55$^\ast$   & 2020 \\ 
		
		\hline
		\textbf{PML} & \textbf{2.16}   & \textbf{-}    \\ 
		\hline
	\end{tabular}%
\end{table}

\textbf{Comparisons on ChaLearn LAP 2015. }We further compared our model with the state-of-the-art models on the ChaLearn LAP 2015. As shown in Table~\ref{tab:ChaLearn}, our method achieves 2.915 MAE which was pretrained on IMDB-WIKI and surpasses the state-of-the-art performance. The results prove that our PML deals with the samples with large variance, while the progressive margin learning achieves to filter noisy instance.

\begin{table}[tb] \small
	\caption{Comparisons of MAEs of our approach compared with different state-of-the-art methods on ChaLearn LAP 2015 dataset.}
	\centering
	\label{tab:ChaLearn}
	\vspace{5pt}
	\begin{tabular}{|c|c|c|c|}
		\hline
		\textbf{Method}  & \textbf{ChaLearn} & \textbf{$\epsilon$-error} & \textbf{Year} \\ 
		\hline
		\hline
		ARN~\cite{DBLP:conf/iccv/AgustssonTG17} & 3.153$^\ast$ & -  & 2017 \\
		TinyAgeNet~\cite{DBLP:conf/ijcai/GaoZWG18} & 3.427$^\dagger$ & 0.301$^\dagger$ & 2018 \\
		CVL\_ETHZ~\cite{2016Deep}  & 3.252$^\ast$ & 0.282$^\ast$ & 2018 \\
		AgeED~\cite{DBLP:journals/pami/TanWLZGL18} & 3.210$^\ast$ & 0.280$^\ast$ & 2018 \\
		ThinAgeNet~\cite{DBLP:conf/ijcai/GaoZWG18} & 3.135$^\dagger$ & 0.272$^\dagger$ & 2018 \\
		ODL~\cite{liu2017ordinal} & 3.950 & 0.312  & 2019 \\
		DHAA~\cite{DBLP:conf/ijcai/TanY0GL19} & \textit{3.052}$^\ast$ & \textit{0.265}$^\ast$ & 2019 \\ 
		\hline
		\textbf{PML} & \textbf{3.455} & \textbf{0.293} & \textbf{-}  \\ 
		\textbf{PML$^\ast$} & \textbf{2.915$^\ast$} & \textbf{0.243$^\ast$} & \textbf{-}  \\ 
		\hline
	\end{tabular}%
\end{table}

\begin{figure}[tb] 
	\begin{center}
		%\fbox{\rule{0pt}{2in} \rule{.9\linewidth}{0pt}}
		\subfigure[Majority Classes]{
			\label{visual.1}
			\centering
			\includegraphics[width=1\linewidth,height=0.55\linewidth]{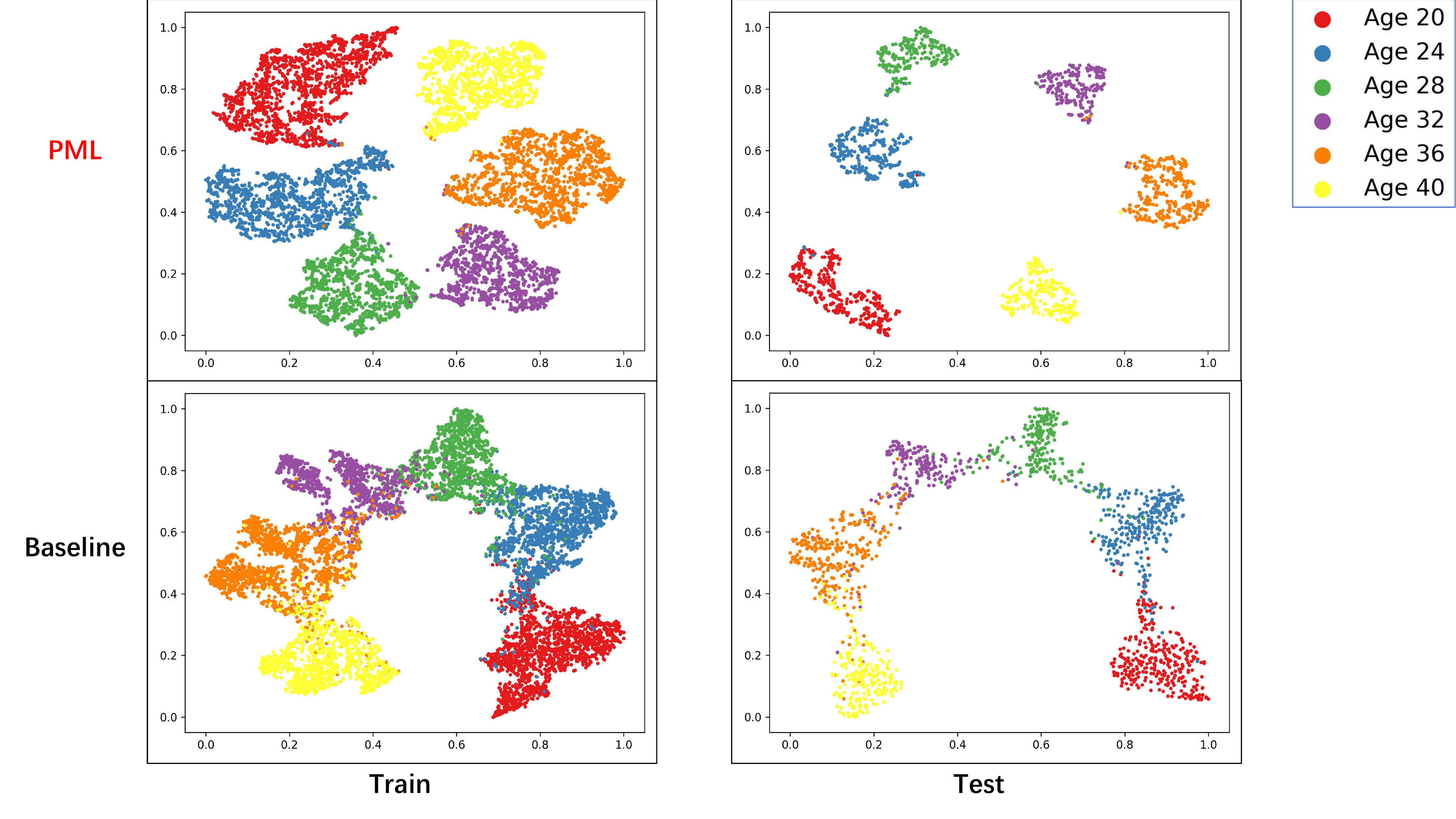}
		}
		
		\subfigure[Minority Classes]{
			\label{visual.2}
			\centering
			\includegraphics[width=1\linewidth,height=0.55\linewidth]{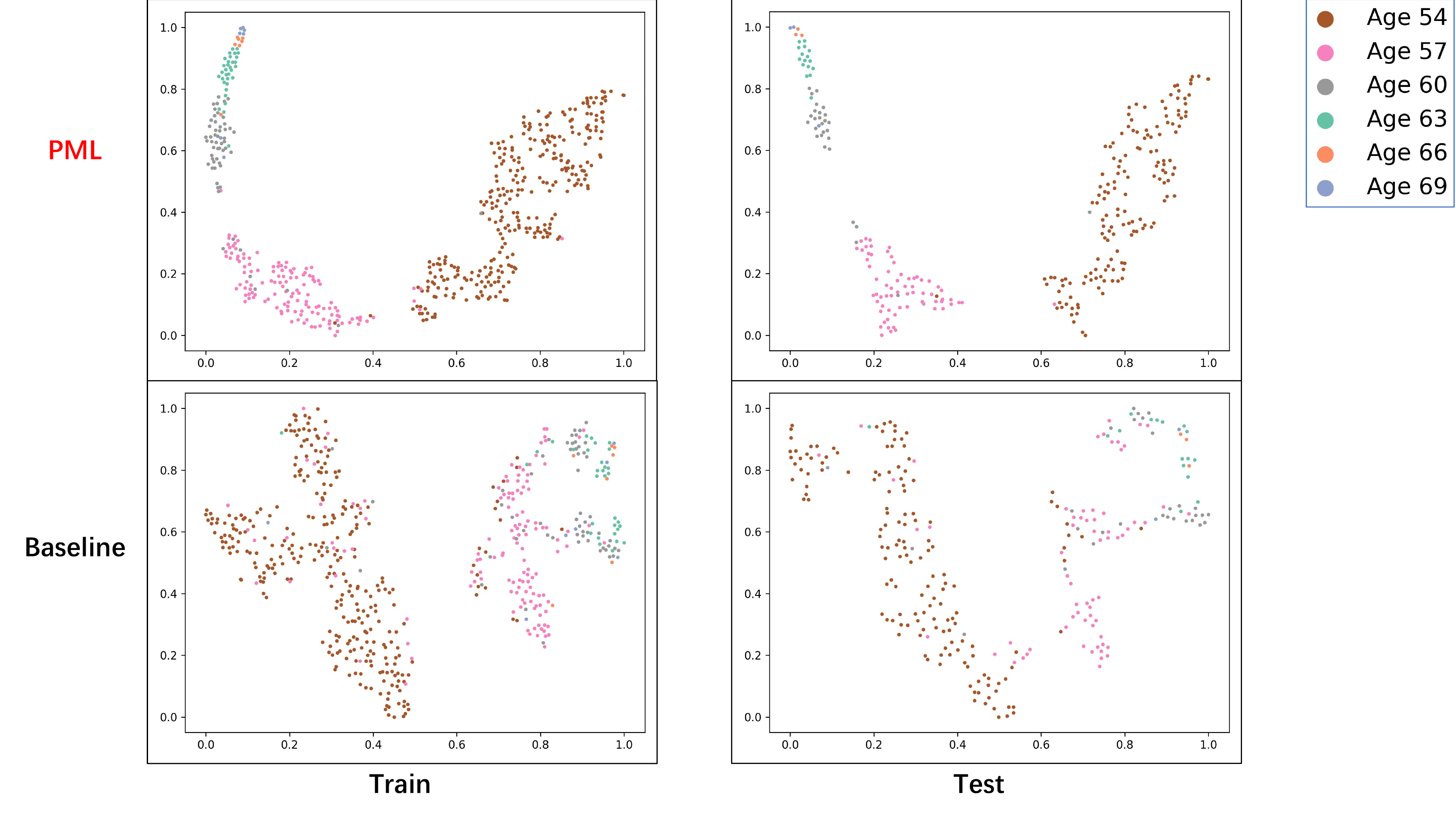}
		}
		
		\subfigure[Majority and Minority Classes]{
			\centering
			\label{visual.3}
			\includegraphics[width=1\linewidth,height=0.55\linewidth]{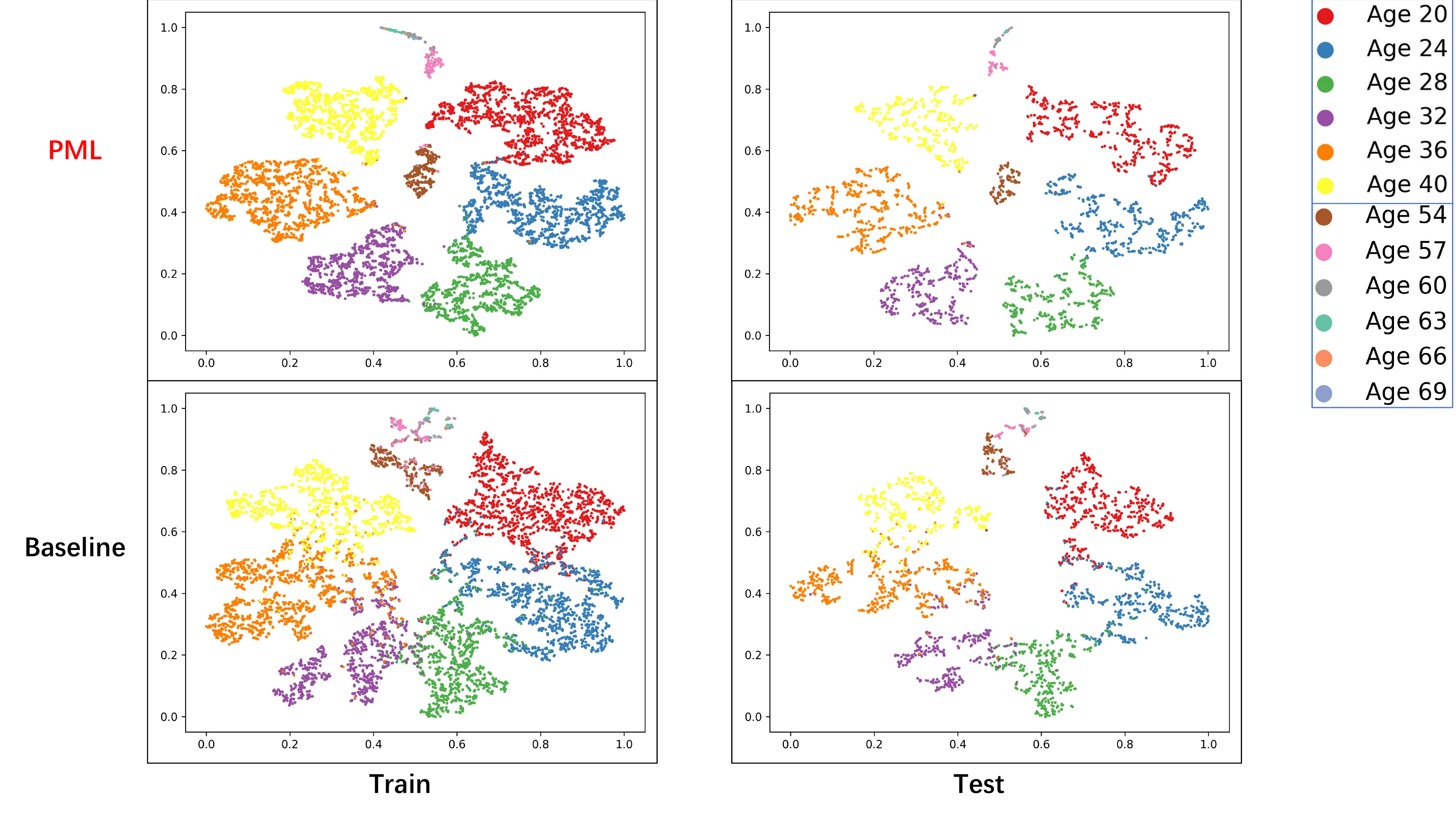}
		}
	\end{center}
	\caption{The visualization of learned feature $\bm{x}$ with t-SNE. We conducted both experiments of training and testing splits on Morph II,  compared with the baseline method without the progressive margin. ~(a)~The visualization of features from 6 majority classes. Seeing from these results, each head class is distinguished by our PML. ~(b)~The visualization of our embedded features from 6 minority classes. With these minority classes, our approach learns more discriminative feature than the baseline. ~(c)~Learned by both the majority and minority classes, the spanned feature space of minority classes is narrowed and disturbed by the majority classes in the baseline method. Fortunately, our PML framework teaches each class to characterize their manifolds by constraining the margin to all classes.~(Zoomed in for better visualization.)}  
	\label{tsne}
\end{figure}

\textbf{Qualitative Results.}
To better demonstrate the effectiveness of our PML intuitively, we visualized the predicted distributions and the learned features with versus without the PML framework. For fair comparisons, we created the baseline model, which has the same architectures as our PML except using the standard KL loss. Fig.~\ref{distribution} shows the six resulting examples from young to old on FG-NET. From the Fig.~\ref{distribution}, we observe that the learned label distributions of our PML significantly suit real-world age correlation than the baseline model. Fig.~\ref{tsne} shows the learned feature with t-SNE~\cite{DBLP:conf/vissym/RauberFT16}. We see that the proposed progressive margin loss effectively guarantees the boundary of each class in the learned embedding space.
\begin{table}[tb] \small
	\caption{Comparisons of MAEs of our approach compared with different state-of-the-art methods on Morph II and ChaLearn LAP 2015 dataset under different curriculum learning protocols.}
	\centering
	\label{tab:cl}
	\vspace{5pt}
	\begin{tabular}{|c|c|c|c|c|}
		\hline
		\textbf{Dataset} & \textbf{Groups} & \textbf{Imbalance Ratio} & \textbf{Sample} & \textbf{MAE} \\ \hline
		Morph II & $\mathcal{D}_1$(20\%) & 27/1 & 1,382 & 3.751 \\
		& $\mathcal{D}_2$(40\%) & 430/1 & 17,611 & 2.828 \\
		& $\mathcal{D}_3$(60\%) & 1054/1 & 37,342 & 2.503 \\
		& $\mathcal{D}_4$(80\%) & 1335/1 & 42,438 & \textbf{2.314} \\ 
		\hline
		\hline
		ChaLearn & $\mathcal{D}_1$(20\%) & 6/1 & 199 & 6.720 \\
		& $\mathcal{D}_2$(40\%) & 16/1 & 840 & 5.306 \\
		& $\mathcal{D}_3$(60\%) & 28/1 & 1,303 & 4.622 \\
		& $\mathcal{D}_4$(80\%) & 60/1 & 2,009 & \textbf{3.878} \\ \hline
		
	\end{tabular}%
\end{table}

\textbf{Analysis. }
To further investigate the effects of our PML regrading with different quantity of training samples, we conducted comparisons on both Morph II and ChaLearn with various courses. For simplicity, we set the dividing line $\{\delta_1, \delta_2, \delta_3, \delta_4\}$ of dataset to $\{20\%, 40\%, 60\%, 80\%\}$ respectively. By following Equ.\ref{eq.curriculum}, a series of curricula from balance to imbalance could be achieved gradually. As the 3\textit{rd} and 4\textit{th} columns of Table~\ref{tab:cl} show, we see that with the quantity of samples increases, the imbalance ratio increases. 
As the Table~\ref{tab:cl} shows, we see that our PML decreases the MAEs from curriculum $\mathcal{D}_1$ to $\mathcal{D}_4$ on both datasets. More specifically, in course $\mathcal{D}_4$, we achieve comparable results with the state-of-the-art methods while training with less samples, \textit{i.e.}, 80\% of the entire dataset. It mainly benefits from the learning instructor of previous curriculum, these instructors assign balanced initial spaces for all classes. Hence, this reduces the probability from trapping into sub-optima.

\section{Conclusions}
In this paper, we have proposed a progressive margin loss framework~(PML)~ for unconstrained facial age classification. The proposed PML has progressively learned the age label pattern by taking both real-world age relations and critical property of the class center into account. 
%Moreover, we have designed a series of curricula to distill the knowledge from previous instructors to reinforce our optimization. 
Experiments on three datasets have demonstrated the effectiveness of proposed approach. In future works, we will focus on self-supervised margin learning in a contrastiveA manner~\cite{DBLP:conf/cvpr/HadsellCL06,DBLP:conf/cvpr/FengXT19} 
by including fewer labels.

{\small
\bibliographystyle{ieee_fullname}
\bibliography{egbib}
}

\end{document}